\documentclass{article}
\usepackage{microtype}
\usepackage{booktabs}
\usepackage{hyperref}

\usepackage[accepted]{icml2024}
\usepackage{amsmath}
\usepackage{amssymb}
\usepackage{mathtools}
\usepackage{amsthm}
\usepackage[capitalize,noabbrev]{cleveref}
\usepackage{subfig}
\usepackage{graphicx}
\newcommand{\mbf}[1]{{\boldsymbol{\mathbf{#1}}}}
\newcommand{\bm}{\mbf}

\theoremstyle{plain}

\theoremstyle{definition}

\theoremstyle{remark}

\usepackage[textsize=tiny]{todonotes}
\usepackage{xspace}
\usepackage{physics}

\usepackage{amsmath,amsfonts,bm}

\def\eqref#1{equation~\ref{#1}}

\def\1{\bm{1}}

\DeclareMathAlphabet{\mathsfit}{\encodingdefault}{\sfdefault}{m}{sl}
\SetMathAlphabet{\mathsfit}{bold}{\encodingdefault}{\sfdefault}{bx}{n}

\newcommand{\E}{\mathbb{E}}

\newcommand{\R}{\mathbb{R}}

\usepackage{hyperref}
\usepackage{url}
\usepackage{graphicx}
\usepackage{physics}
\usepackage{booktabs}
\usepackage{colortbl}
\usepackage{fixltx2e}
\usepackage{xspace}
\usepackage{algorithm}
\usepackage{algorithmic}
\usepackage{subfig}
\usepackage{makecell}
\usepackage{mathtools}
\usepackage{wrapfig}

\renewcommand{\E}[2]{\mathbb{E}_{#1}\qty[#2]}

\newcommand{\N}{\mathcal{N}}

\newcommand{\rebut}[1]{\textcolor{black}{#1}}
\newcommand{\method}{AFT}
\renewcommand{\R}{\mathbb{R}}
\newcommand{\din}{d_\mathrm{in}}
\newcommand{\dout}{d_\mathrm{out}}
\newcommand{\dphi}{d_\phi}
\newcommand{\dpsi}{d_\psi}

\icmltitlerunning{Transferring Knowledge from Large Foundation Models to Small Downstream Models}

\begin{document}

\twocolumn[

\icmltitle{Transferring Knowledge from \\ Large Foundation Models to Small Downstream Models}

\icmlsetsymbol{equal}{*}

\begin{icmlauthorlist}
\icmlauthor{Shikai Qiu}{aws,nyu,intern}
\icmlauthor{Boran Han}{aws}
\icmlauthor{Danielle C. Maddix}{aws}
\icmlauthor{Shuai Zhang}{aws}
\icmlauthor{Yuyang Wang}{aws}
\icmlauthor{Andrew Gordon Wilson}{aws,nyu}
\end{icmlauthorlist}

\icmlaffiliation{nyu}{Department of Computer Science, New
York University, NYC, USA}
\icmlaffiliation{aws}{AWS AI Labs, Santa Clara, CA, USA}
\icmlaffiliation{intern}{Work done during an internship at AWS}

\icmlcorrespondingauthor{Shikai Qiu}{sq2129@nyu.edu}
\icmlcorrespondingauthor{Boran Han}{boranhan@amazon.com}
\icmlcorrespondingauthor{Andrew Gordon Wilson}{andrewgw@cims.nyu.edu}

\vskip 0.3in
]

\printAffiliationsAndNotice{}

\begin{abstract}
How do we transfer the relevant knowledge from ever larger foundation models into small, task-specific downstream models that can run at much lower costs? Standard transfer learning using pre-trained weights as the initialization transfers limited information and commits us to often massive pre-trained architectures.
 This procedure also precludes combining multiple pre-trained models that learn complementary information. To address these shortcomings, we introduce \emph{Adaptive Feature Transfer} (AFT). Instead of transferring weights, AFT operates purely on features, thereby decoupling the choice of the pre-trained model from the smaller downstream model. Rather than indiscriminately compressing all pre-trained features, AFT adaptively transfers pre-trained features that are most useful for performing the downstream task, using a simple regularization that adds minimal overhead. Across multiple vision, language, and multi-modal datasets, AFT achieves significantly better downstream performance compared to alternatives with a similar computational cost. Furthermore, AFT reliably translates improvement in pre-trained models into improvement in downstream performance, even if the downstream model is over $50\times$ smaller, and can effectively transfer complementary information learned by multiple pre-trained models.
\end{abstract}

\vspace{-5mm}
\section{Introduction}

\begin{figure*}[!t]
\centering
    \subfloat[Information diagram for \method]{
    \includegraphics[height=0.22\linewidth]{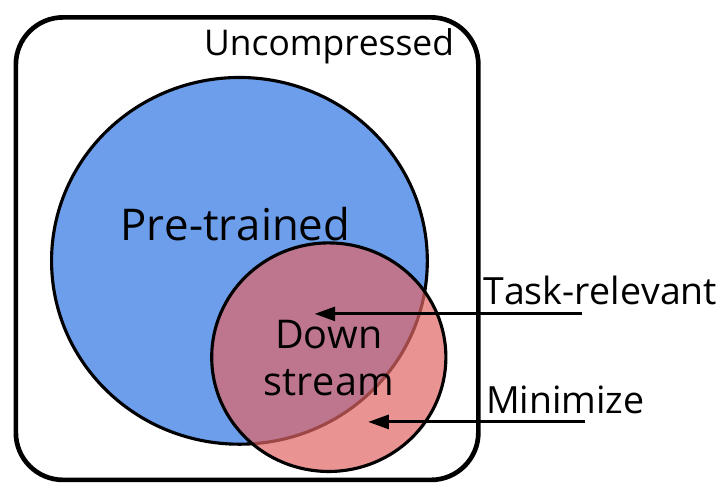}
    \label{fig:illustration}
    }
    \hspace{-3mm}
    \subfloat[Aggregated performance]{
    \includegraphics[height=0.22\linewidth]{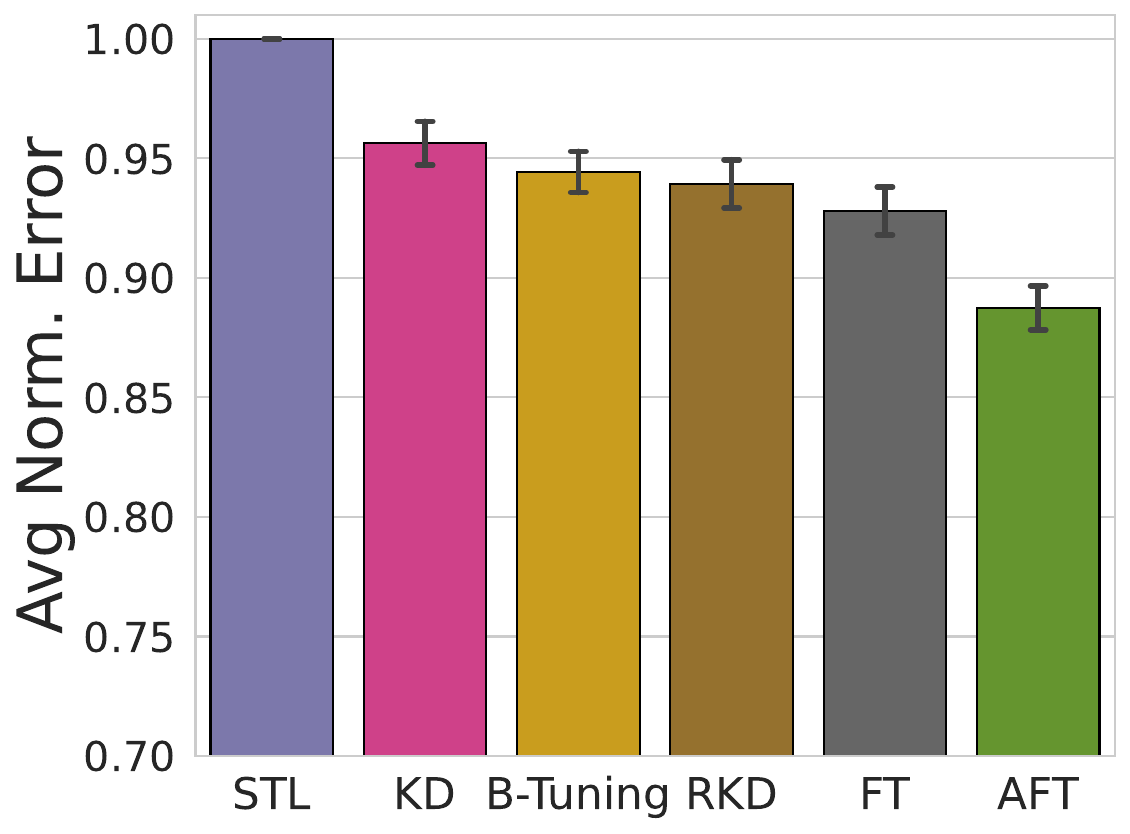}
    }
    \hspace{1mm}
    \subfloat[Using stronger pre-trained models]{
    \includegraphics[height=0.22\linewidth]{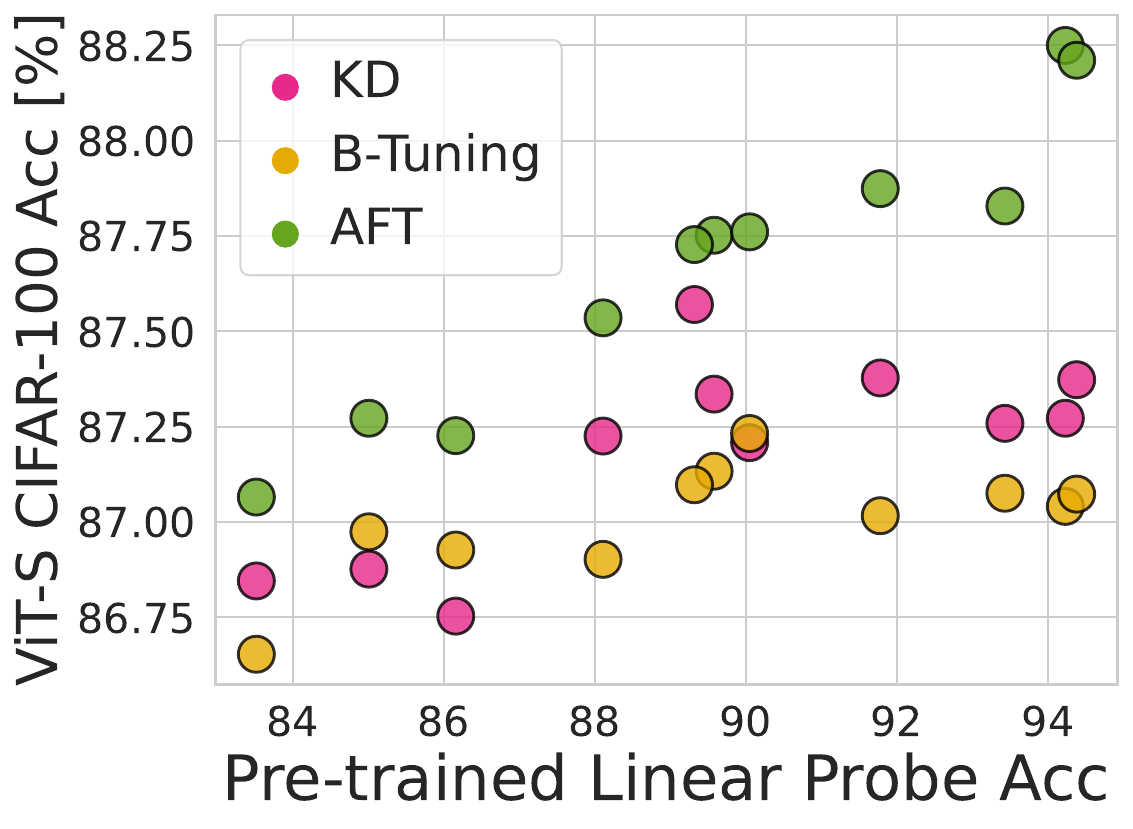}
    }
\caption{\textbf{Adaptive Feature Transfer (\method) transfers knowledge from large foundation models into small downstream models, improving downstream performance with minimal cost}. (\textbf{a}) AFT regularizes the downstream model to prioritize learning the task-relevant subset of pre-trained features ($\mathrm{blue} \cap \mathrm{red}$) over entirely new features ($\mathrm{red} \setminus \mathrm{blue}$). The blue region represents information in pre-trained features, red represents information in downstream features, and inside the square boundary represents all information in the raw, uncompressed input. (\textbf{b}) Over 6 vision datasets and 8 NLP datasets, \method\ significantly outperforms standard transfer learning (STL), knowledge distillation (KD)~\citep{hinton2015distilling,romero2014fitnets}, including its more sophisticated variants relational knowledge distillation (RKD) \citep{park2019relational} and factor transfer (FT) \citep{kim2018paraphrasing}, and B-Tuning~\citep{you2022ranking}. Error is normalized by STL error and averaged over datasets and downstream models, including ViT-S, MLP Mixer-B, ResNet-50, BERT-S, and DistillBERT. Error bars show standard errors across models and datasets. (\textbf{c}) \method\ is the most effective at translating improvements in pre-trained models to improvements in downstream performance. See \Cref{sec:experiments} for experiment details.}
\label{fig:top-fig}
    \vspace{-4mm}
\end{figure*} 
Despite the growing importance of transfer learning, it remains standard practice to simply start with some pre-trained weights as an initialization for fine-tuning on downstream data. This procedure only transfers generic and limited information and the computational burden of fine-tuning and deploying pre-trained models is quickly becoming prohibitive with increases in model size ~\citep{foundation_models, brown2020language, ViT, zhai2022scaling}.
Furthermore, this approach precludes transferring from multiple pre-trained models that learn complementary information due to different pre-training strategies, when a variety of distinctly pre-trained models have become available, especially in domains like computer vision~\citep{oquab2023dinov2, clip, kolesnikov2020big, chen2020simple}.

In principle, however, this transfer from large foundation models to small downstream models should not only be possible but also natural, since the downstream models need not indiscriminately compress all knowledge learned by pre-training, but only inherit the task-revelant knowledge. Leveraging this insight, we propose \textit{Adaptive Feature Transfer} (\method), illustrated in \Cref{fig:illustration}, a simple, general, and efficient method to adaptively transfer task-relevant knowledge from a set of pre-trained models into a small downstream model, with negligible cost compared to standard training. Viewing pre-trained features as a compressed representation of the input containing highly relevant information for downstream predictions, \method\ steers the downstream model to prioritize learning the task-relevant subset of pre-trained features over entirely new features representing information about the raw input but not preserved by pre-training. Crucially, recognizing not all pre-trained features are relevant for a specific downstream task, \method\ 
discourages the downstream model from learning irrelevant features.

Across multiple vision, language, and multi-modal datasets, we show \method\ delivers a substantial performance improvement when transferring from some of the strongest open-source vision and language foundation models, compared to alternatives with a similar computational cost: direct fine-tuning of the downstream model with standard transfer learning, B-Tuning~\citep{you2022ranking}, an efficient method multi-source and cross-architecture transfer learning, and knowledge distillation from the pre-trained to the downstream model~\cite{hinton2015distilling,romero2014fitnets,park2019relational,kim2018paraphrasing}. 
Moreover, we find \method\ is particularly effective at translating improvements in pre-trained models into improvements in downstream performance (Figure \ref{fig:top-fig}). Our code is available at \url{https://github.com/amazon-science/adaptive-feature-transfer}.

\section{Related Work}
\label{sec:bg}
We review the standard transfer learning approach and methods that enable efficient transfer learning from multiple sources and across architectures.
\vspace{-4mm}
\paragraph{Transfer learning.}
Standard transfer learning (STL) proceeds by loading a pre-trained parameter vector as the initialization for parameters $\theta$ of a downstream model with the same architecture, followed by updating $\theta$ by minimizing the downstream loss $L(\theta)$, known as fine-tuning \citep{survey_transfer_learning}. This simple approach has enabled state-of-the-art performances on a wide range of vision \citep{ViT, oquab2023dinov2, he2015deep} and language tasks \citep{Bert, touvron2023llama}.  

\citet{shwartz2022pre} note that STL merely transfers an initialization, and that our knowledge of the source task should affect the shapes and locations of optima on the downstream task. To transfer additional information, \citet{shwartz2022pre} propose a Bayesian transfer learning approach by regularizing the downstream model with a Gaussian prior centered at the pre-trained weights, with a covariance matrix such that $\theta$ is allowed large variance in directions where pre-training loss increases slowly.

\vspace{-4mm}
\paragraph{Efficient multi-source transfer learning.}
To transfer from multiple sources without fine-tuning many pre-trained models, \citet{lee2019learning} propose to learn a classifier defined as a weighted combination of frozen pre-trained features, where the weights are derived from non-linear maximal correlation analysis. \citet{chang2022towards} uses a mixture-of-experts model to combine complementary information across different models and datasets in material sciences. \citet{shu2021zoo} develops Zoo-Tuning to aggregate the parameters from multiple pre-trained models into a single downstream model, all assumed to have the same architecture. In addition, several works propose to rank and select in advance a subset of pre-trained models or features for transferring to a specific downstream task~\citep{you2022ranking,fumero2023leveraging, deshpande2021linearized}, thus reducing the cost of exploration when a large number of pre-trained models are available. As these methods still reuse the pre-trained architecture for the downstream task, they are only useful for reducing the cost of training, but not the cost of deploying large pre-trained architectures. Moreover, methods such as Zoo-Tuning cannot be applied to transfer across architectures, limiting the choice of pre-trained models.

\vspace{-4mm}
\paragraph{Cross-architecture transfer learning.}
B-Tuning \citep{you2022ranking} is a recently proposed method that enables cross-architecture transfer by regularizing the downstream model with a prior defined by the approximate posterior of a linear model conditioned on pre-trained features. Unlike the prior in \citet{shwartz2022pre}, this prior is defined in function space rather than parameter space, and can therefore be used for downstream models of any architecture. On transferring from multiple pre-trained vision models, \citet{you2022ranking} shows B-Tuning outperforms both knowledge distillation and Zoo-Tuning. 

An alternative approach to cross-architecture transfer is knowledge distillation (KD)~\citep{hinton2015distilling}. While the original KD trains the student to perform the same task as the teacher, feature-based KD can be applied to transfer the knowledge learned by a teacher pre-trained on a different but related task to a downstream student model, by training it to predict the teacher's features rather than logits \citep{romero2014fitnets, 9009585, huang2017like, 10.1609/aaai.v33i01.33013779, gu2023knowledge, yim2017gift, ahn2019variational, you2022ranking}. In this approach, the student is usually trained to minimize a regression objective $\E{x}{\norm{\phi_T(x) - V\phi_S(x)}^2_2},$ where $\phi_S$ and $\phi_T$ denote the student and teacher features, and $V$ is a learned transformation that can account for the difference in dimensionality and the arbitrariness of the choice of coordinates. Many works have proposed more sophisticated version of feature-based KD, such as relational knowledge distillation (RKD) \citep{park2019relational} that aims to capture the relation between the features of different inputs rather than their absolute values, and factor transfer \citep{kim2018paraphrasing}, which trains the student to predict a compressed version of the teacher features learned through an autoencoder. Other works, such as \citet{jang2019learning, ji2021show}, focus on incorporating features from many intermediate layers.

\paragraph{Difference between AFT and prior works.}
As we shall explain in detail in \Cref{sec:method}, AFT is conceptually distinct from B-Tuning and KD, though they all use pre-trained features to regularize the downstream model.
The main difference between our approach and B-Tuning is that 1) we regularize the downstream model's features rather than predictions, which allows more information to be transferred into the downstream model (features are often higher dimensional than the outputs), and 2) we learn the importance of each pre-trained feature during training on the downstream task rather than determining it ahead of time based purely on the posterior predictive mean of pre-trained models, which fails to take into account any property of the downstream model. In contrast to KD, AFT does not penalize the downstream model (student) from forgetting some of the pre-trained (teacher) features, and only penalizes learning extra features not extracted from pre-training.

\section{Adaptive Feature Transfer}
\label{sec:method}

We now introduce Adaptive Feature Transfer (\method), a method that adaptively transfers task-relevant knowledge from large foundation models to a small downstream model with negligible overhead compared to standard training.

\subsection{An informative prior from pre-trained features} 

The core intuition behind \method\ is that we want the downstream model to prefer making predictions based on information already present in the pre-trained features, as they are highly likely to contain useful knowledge for the downstream task, but without necessarily using all pre-trained features, since not all of them will be relevant to the downstream task. We now formalize this simple intuition mathematically by defining a prior for downstream learning. Let $\theta \in \R^P$ be the downstream model parameters, the random variable $X \in \R^{\din}$ be the downstream inputs, $\Phi = \phi_\theta(X) \in \R^{\dphi}$ be the features of the downstream model, $Y = W \Phi  \in \R^{\dout}$ be the downstream model outputs, and $\Psi=\psi(X) \in \R^{\dpsi}$ be a list of frozen pre-trained features, formed by concatenating the last layer features from an arbitrary number of pre-trained models. To encourage the desired behavior, we define a prior that favors low mutual information between downstream features $\Phi$ and the input $X$ conditioned on the pre-trianed features $\Psi$,
\begin{equation}
    p(\theta) \propto \exp(- \beta I(\Phi;X|\Psi)),
\end{equation}
where the $I(\Phi;X|\Psi)$ measures the amount of information about $X$ encoded in downstream features $\Phi$ but not in the pre-trained features $\Psi,$ visualized in \Cref{fig:top-fig} as the area of $\mathrm{red} \setminus \mathrm{blue}$, and $\beta > 0$ controls the strength of this prior. The mutual information is given by
\begin{align}
        I(\Phi; X | \Psi) &= H(\Phi|\Psi) - H(\Phi|X, \Psi) \\
        &= \E{\Phi, \Psi}{-\log p(\Phi|\Psi)} + c \\
        &\leq \min_\mu \E{\Phi, \Psi}{-\log q_\mu(\Phi|\Psi)} + c,
\end{align}
where $H$ denotes the conditional entropy. $H(\Phi|X, \Psi)$ is some constant $c$ since $\Phi$ is deterministic given $X$ and we used a variational distribution  $q_\mu(\Phi|\Psi)$ with variational parameters $\mu$ to approximate the inaccessible conditional density $p(\Phi|\Psi)$ and thus bound the mutual information. 

To train the downstream model, we seek the most likely parameters conditioned on the data under this prior, by minimizing the bound on the negative log posterior, equal to $L(\theta) + \beta R(\theta)$, where $L(\theta)$ is the unregularized loss (e.g. cross-entropy loss) and $R(\theta)$ is the bound on the mutual information given by
\begin{align}
    R(\theta) = \min_\mu \E{\Phi, \Psi}{-\log q_\mu(\Phi|\Psi)},
\end{align}

where the expectation can only be estimated using training samples. The effect of optimizing this objective is to maximize the downstream data fit while minimizing the information in downstream features $\Phi$ that cannot be decoded from the pre-trained features $\Psi$ via the map $q_\mu(\Phi|\Psi),$ after optimizing for variational parameters $\mu$. We consider a simple Gaussian parameterization $q_\mu(\Phi|\Psi) = \N(\Phi|\mu \Psi, I)$, where $\mu: \R^{\dpsi} \to \R^{\dphi}$ is an affine transformation, which leads to:
\begin{align}
    \label{eq:l2-reg}
    R(\theta) = \min_{\mu}\E{\Phi, \Psi}{\norm{\Phi - \mu\Psi}^2},
\end{align}
after ignoring some $\theta-$independent constants. Since the minimization over the offsets in the affine transformation is equivalent to subtracting the mean from both $\Phi$ and $\Psi,$ we will henceforth assume that $\Phi$ and $\Psi$ have been pre-processed to have zero-mean and assume $\mu \in \R^{\dphi \times \dpsi}$ to be a linear transformation. 

By comparison, the KD objective is equivalent to
\begin{align}
\label{eq:kd}
R_\mathrm{KD}(\theta) = \min_{V}\E{\Phi, \Psi}{\norm{V\Phi - \Psi}^2},
\end{align}
with $V \in \R^{\dpsi \times \dphi}$. The regularization we introduce moves the learnable transformation to act on the pre-trained features instead of the downstream features. This simple modification makes the objective more suitable for transfer learning.
While minimizing the KD objective requires the downstream $\Phi$ features to contain all information needed to predict the pre-trained features $\Psi$, even if some are irrelevant or harmful to the downstream task, our objective $R(\theta)$ only requires the downstream features $\Phi$ to lie in the span of the pre-trained features $\Psi$, allowing $\Phi$ to encode only a subset of information in $\Psi$.
With this simple but significant change to the knowledge distillation objective, we incentivize an adaptive transfer of pre-trained features to the downstream task. As we will show, this objective leads to significant performance gains for transfer learning with almost no additional cost and is particularly effective at translating improvements in pre-trained models to downstream performance.

\subsection{Improving the objective using kernels}
While conceptually straightforward, evaluating and minimizing the regularization $R(\theta)$ in Eq.~\ref{eq:l2-reg} introduces both optimization and statistical challenges: 1) since evaluating $R(\theta)$ requires finding the optimal variational parameters $\mu$, which changes every time we update $\theta$, we want to simplify the optimization problem for $\mu$ to minimize its computational overhead, and 2) since we wish to estimate the true $R(\theta)$ whose exact value is given by an expectation over the true rather than empirical distribution of $\Phi$ and $\Psi,$ we want to avoid over-fitting to the training data when optimizing for $\mu$ once we replace the expectation in Eq.~\ref{eq:l2-reg}  with its empirical estimate, especially since transfer learning often involves small downstream datasets.

We now show how to exploit a kernel formulation of the objective to further mitigate both challenges. Recall that the behavior of a linear model $f(\cdot) = w^\top \phi(\cdot)$ is completely characterized by its kernel $k_\Phi(x, x') = \phi(x)^\top\phi(\rebut{x'})$. From a kernel perspective, the existence of $\mu \in \R^{\dphi \times \dpsi}$ such that $\Phi = \mu \Psi$ is equivalent to the existence of $\tilde{\mu} \in \R^{\dphi \times \dpsi}$ such that $k_\Phi = k_{\tilde{\mu}\Psi}.$ Therefore, we replace the $\ell_2$ distance between the features with a distance between their kernel functions,
\begin{align}
    \label{eq:kernel-reg}
    R_\mathrm{AFT}(\theta) = \min_{\mu} \sqrt{\E{}{\qty(k_\Phi(X, X') - k_{\mu\Psi}(X, X'))^2}},
\end{align}
where $X$ and $X'$ are drawn from the input distribution. As with the previous objective in Eq.~\ref{eq:l2-reg}, this objective achieves a minimum value of 0 if and only if each $\phi_i(\cdot), i = 1, ..., \dphi,$ is in the span of $\{\psi_i(\cdot)\}_{i=1}^{\dpsi}.$ However, the kernel formulation has the key advantage that part of the optimization problem over $\mu$ is done automatically since the kernel is invariant under any orthogonal transformation of the features, implying that we only need to optimize $\mu$ up to an orthogonal transformation, significantly reducing the complexity of the inner optimization. This reduction of complexity simply reflects the fact there is no substantive difference between two models whose features only differ by an orthogonal transformation, e.g. a permutation or rotation of the feature dimensions.

To prevent over-fitting the variational parameters $\mu$ to the empirical distribution of the features, we parameterize $\mu$ as a diagonal matrix $\mathrm{diag}(\sigma(s)),$ \rebut{i.e. $\mu_{ii} = \sigma(s_i)$}, where $\sigma$ is the sigmoid function and $s$ is a $\dpsi$-dimensional vector. Doing so greatly reduces the number of variational parameters to optimize, while retaining the ability for the model to weigh each dimension of the pre-trained features differently. Note that choosing a diagonal $\mu$ is always admissible in the kernel formulation, which does not require the features to have the same dimensions. Furthermore, due to the invariance of the kernel under orthogonal transformations, we are effectively searching over all $\mu' = U\mu = U \mathrm{diag}(s) \in \R^{\dpsi \times \dpsi},$ where $U \in \R^{\dpsi \times \dpsi}$ is any orthogonal matrix, without actually optimizing the dense matrix $U$ which has significantly more parameters than $\mu$. Finally, we normalize the features to have unit $\ell_2$ norm before computing the respective kernels, i.e., $k_\Phi(x,x') \coloneqq \phi(x)^\top\phi(x')/\norm{\phi(x)}\norm{\phi(x')},$ to reduce the variance in the kernel entries. 

\begin{figure}[!t]
\centering
    \subfloat[DINOv2 ViT-G/14 to ViT-S]{
    \includegraphics[width=0.45\linewidth]{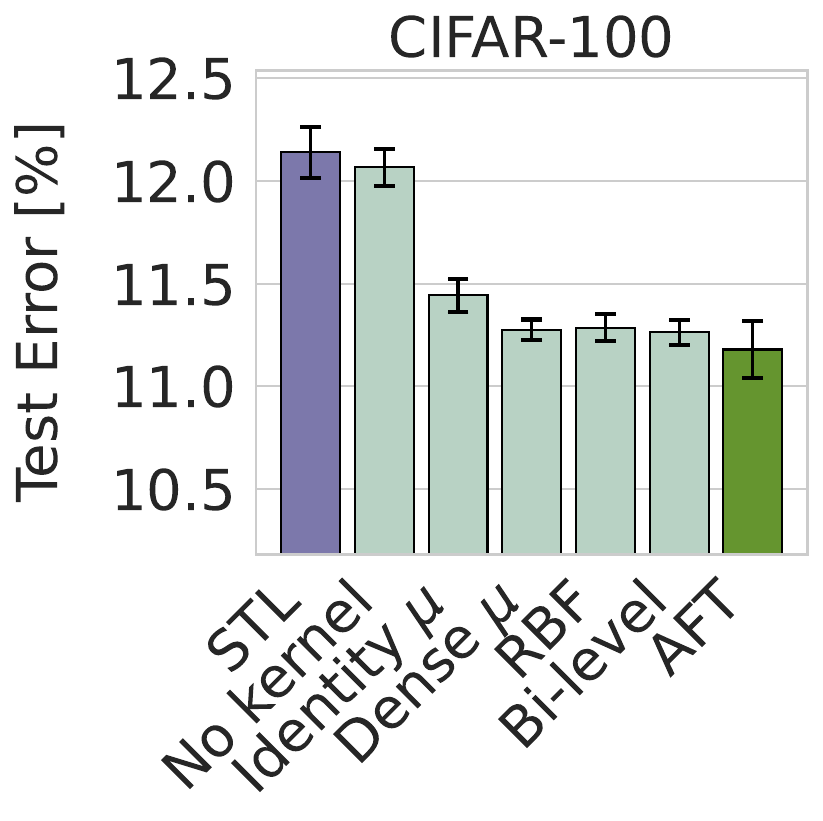}
    }
    \subfloat[Flan-T5 Large to BERT-S]{
    \includegraphics[width=0.45\linewidth]{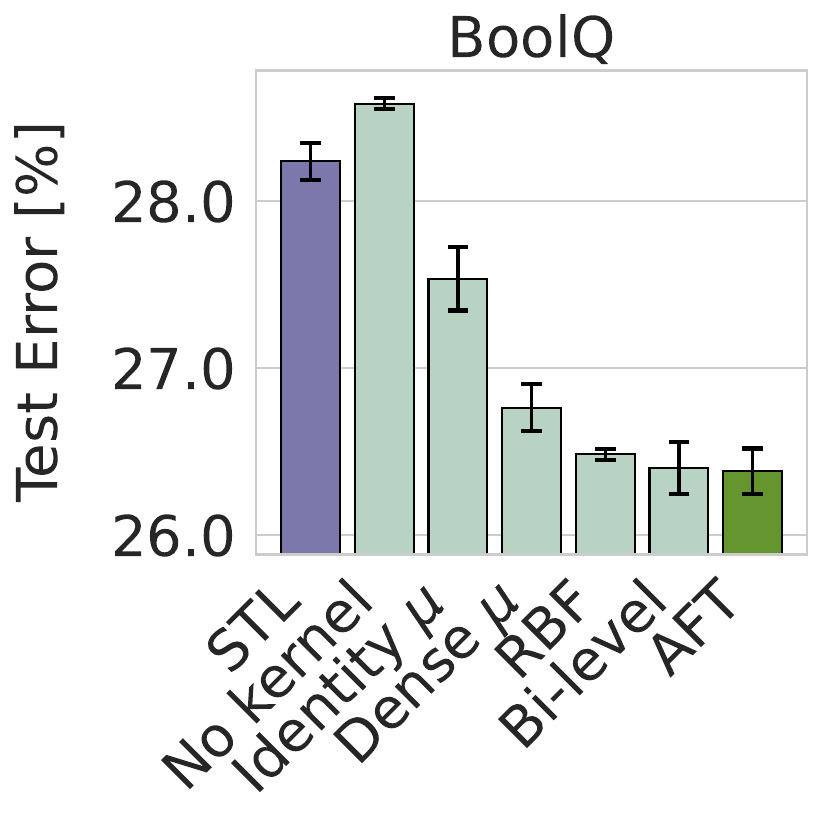}
    }
    \label{fig:ablation}
   \caption{\rebut{\textbf{Ablation experiments}. Using the kernel and learning $\mu$ is essential for \method's performance, whereas using an RBF kernel and bi-level optimization over $(\mu, \theta)$ barely impacts performance. Making $\mu$ dense slightly hurts performance.}}
\end{figure}

In Section~\ref{sec:ablation}, we compare \method\ with its other variants and show that both using the kernel formulation and learning a diagonal $\mu$ are essential to its performance (\Cref{fig:ablation}). We also verify that the learned $\mu$ indeed places higher weights on more informative features (\Cref{fig:noise_rho}), allowing AFT to achieve robust performance even when a significant fraction of the pre-trained features is noise (\Cref{fig:noise_acc}).

\begin{algorithm}[!t]
  \caption{Adaptive Feature Transfer (\method)}
  \label{alg:dft}
  \begin{algorithmic}[1]
  \small
    \REQUIRE Pre-computed pre-trained features, downstream data, downstream model $f_\theta = W \circ \phi_\theta,$ \\\rebut{downstream loss function $L,$} batch size $B,$ learning rates $ (\eta_1, \eta_2) $, regularization coefficient $ \beta $
      \FOR{each mini-batch $ X_{\mathrm{batch}} \in \mathbb{R}^{B \times \din}, Y_{\mathrm{batch}} \in \mathbb{R}^{B \times \dout}, \Psi_{\mathrm{batch}} \in \mathbb{R}^{B \times \dpsi} $}
        \STATE Compute features $\Phi_{\mathrm{batch}} = \phi_\theta(X_{\mathrm{batch}}) \in \mathbb{R}^{B \times \dphi}$ and outputs $\hat{Y}_\mathrm{batch} = \Phi_{\mathrm{batch}} W^\top$
        \STATE Scale pre-trained features $\Psi_{\mathrm{batch}} \leftarrow \Psi_{\mathrm{batch}} \mu^\top$
        \STATE Subtract the mini-batch mean from $\Phi_{\mathrm{batch}}$ and $\Psi_{\mathrm{batch}}$ and normalize each row
        \STATE Compute $B \times B$ mini-batch kernels $K^\Phi_{\mathrm{batch}} = \Phi_{\mathrm{batch}}\Phi_{\mathrm{batch}}^\top, K^{\mu\Psi}_{\mathrm{batch}} = \Psi_{\mathrm{batch}}\Psi_{\mathrm{batch}}^\top$
        \STATE Compute mini-batch loss $\rebut{\hat{L}(\theta) = L(\theta, Y_{\mathrm{batch}}, \hat{Y}_{\mathrm{batch}})}$ and the kernel distance estimate:
        \vspace{-0.2cm}
        \[ \hat{\delta}(\theta, \mu) = \frac{1}{B} \norm{K^\Phi_{\mathrm{batch}} - K^{\mu\Psi}_{\mathrm{batch}}}_F \]
        \vspace{-0.4cm}
        \STATE Update $ \theta $ and $\mu$: 
        \[ \theta \leftarrow \theta - \eta_1 \nabla_{\theta} \qty(\hat{L}(\theta) + \beta \hat{\delta}(\theta, \mu)), \quad\mu \leftarrow \mu - \eta_2 \nabla_{\mu} \hat{\delta}(\theta, \mu) \]
      \ENDFOR
  \end{algorithmic}
\end{algorithm}

\paragraph{Stochastic kernel distance estimation.}
For an efficient implementation, we estimate the kernel distance $\sqrt{\E{}{\qty(k_\Phi(X, X') - k_{\mu\Psi}(X, X'))^2}}$ with a mini-batch estimate $\sqrt{\frac{1}{B^2} \sum_{i=1}^{B}\sum_{j=1}^{B} \qty(k_\Phi(x_i, x_j) - k_{\mu\Phi}(x_i, x_j))^2} = \frac{1}{B}\norm{K^\Phi_{\mathrm{batch}} - K^{\mu\Psi}_{\mathrm{batch}}}_F,$ where $K^{\Phi}_{\mathrm{batch}}$ and $K^{\mu\Psi}_{\mathrm{batch}}$ are kernel matrices evaluated on a batch of $B$ inputs. We then perform gradient-descent over $(\theta, \mu)$ jointly. Algorithm~\ref{alg:dft} details the training procedure, simplifying the update expression assuming SGD. 

\paragraph{Negligible training overhead.}
We compute and cache the pre-trained features on the training set once and simply retrieve them during training without spending additional time to compute them. \Cref{tab:train_overhead} compares the runtime on an NVIDIA A100 GPU for training ViT-S/16 (22M parameters) for one epoch on CIFAR-100 using STL and AFT, where AFT uses pre-trained features from OpenCLIP ViT-L/14 (303M parameters) \citep{cherti2023reproducible}. As expected, the overhead of retrieving pre-computed features and computing the kernel distance is negligible compared to standard training. Pre-computing the features incurs only a one-time cost, which takes about 9 minutes for OpenCLIP ViT-L/14 on the CIFAR-100 training set.
 
\begin{table}[b!]
\small
\centering
\caption{AFT has negligible training overhead compared to standard transfer learning. We report 1 epoch training time on CIFAR-100 for ViT-S/16 with STL and AFT, where AFT transfers features from OpenCLIP ViT-L/14.}
\setlength{\tabcolsep}{6pt}
\renewcommand{\arraystretch}{1.2}
\begin{tabular}{@{}llll@{}}
\toprule
Method & Pre-trained ($\psi$) & Downstream ($\phi$) & Time (min) \\
\midrule
STL    & N/A                 & ViT-S/16            & $1.74$       \\
AFT    & OpenCLIP ViT-L/14   & ViT-S/16            & $1.77$      \\
\bottomrule
\end{tabular}
\label{tab:train_overhead}
\end{table}

\begin{figure*}[!t]
\centering
    \includegraphics[width=0.7\linewidth]{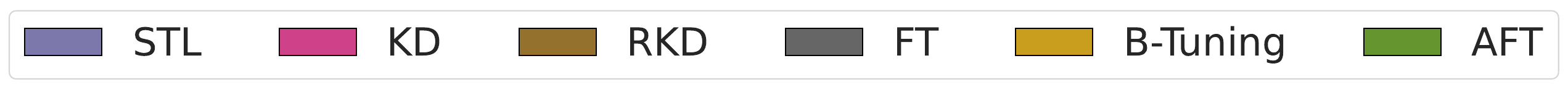}
    \subfloat[Aggregated error]{
    \includegraphics[height=0.2\linewidth]{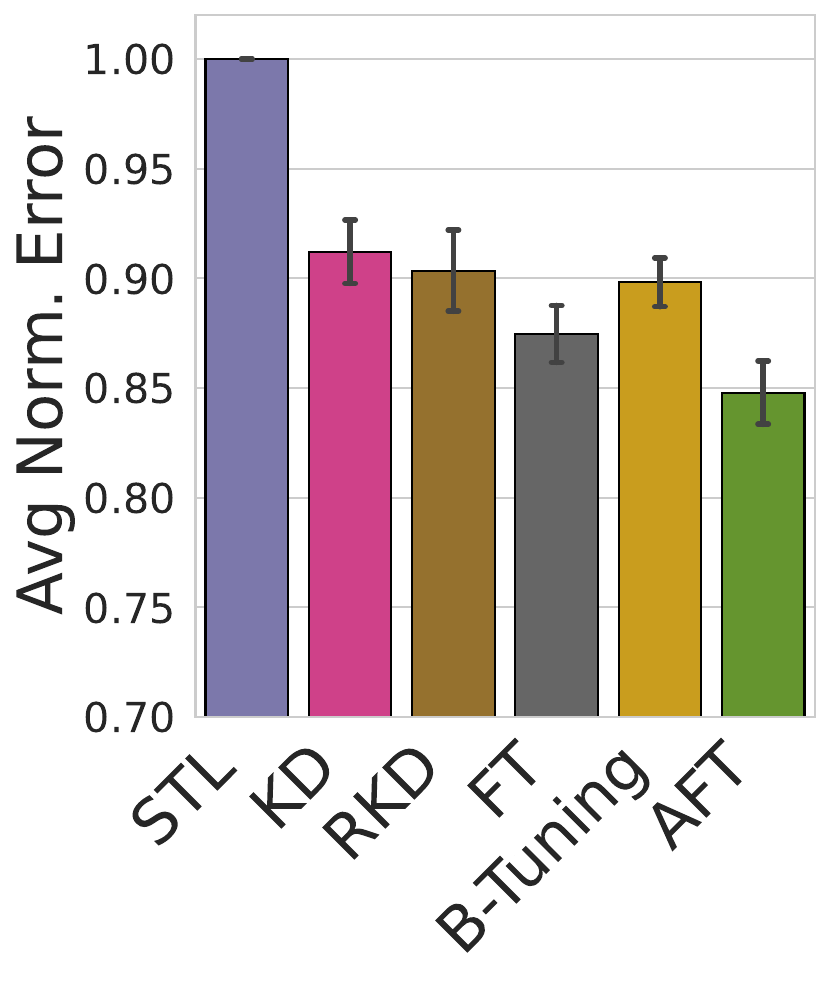}    
    \label{fig:vision_agg}
    }
    \subfloat[Error across models and datasets]{
    \includegraphics[height=0.2\linewidth]{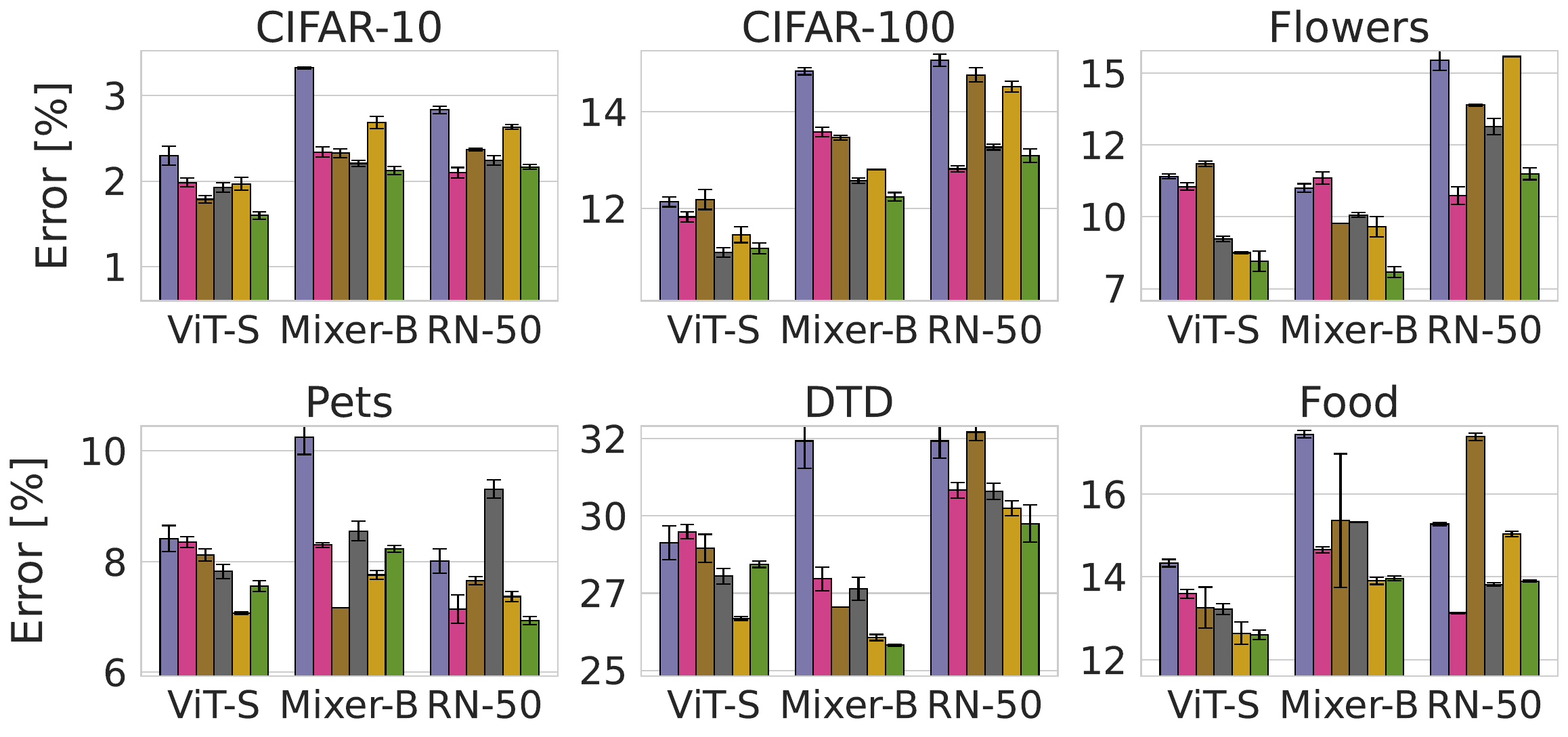}
    \label{fig:vision_all_datasets}
    }
    \subfloat[ViT-S]{
    \includegraphics[height=0.2\linewidth]{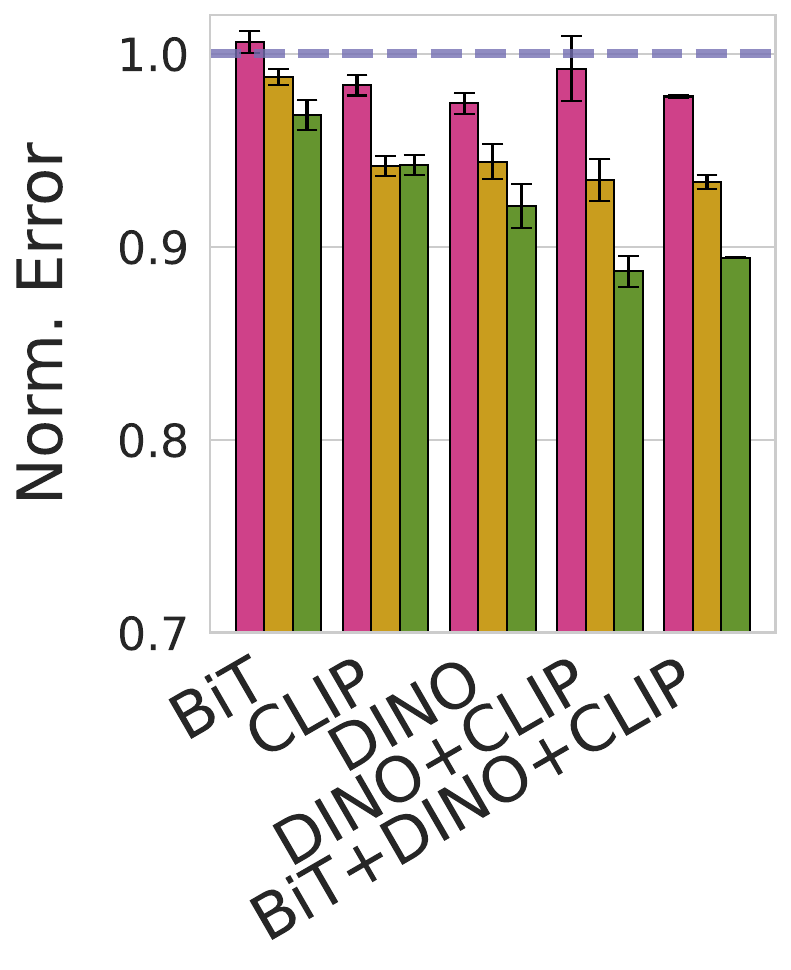}
    \label{fig:vision_combine_vit}
    }
    \subfloat[Mixer-B]{
    \includegraphics[height=0.2\linewidth]{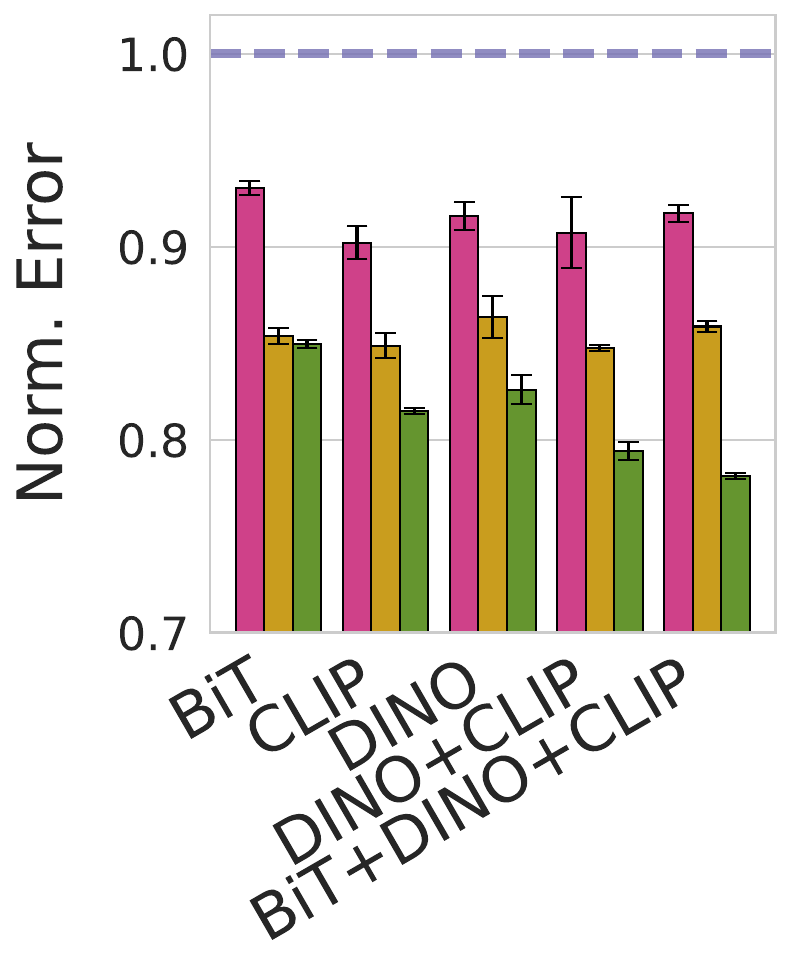}
    \label{fig:vision_combine_mixer}
    }
   \caption{\textbf{Evaluation on 6 vision datasets using ViT-S, MLP-Mixer-B, and ResNet-50 as downstream models}. (\textbf{a}) \method\ achieves the lowest normalized error, averaged across all 6 datasets, 3 downstream models, and 3 seeds when transferring from DINOv2 ViT-G/14. The error is normalized by the STL error before averaging. Error bars show standard errors of the aggregated performance. (\textbf{b}) Breakdown of unnormalized error for each downstream model and dataset. Error bars show standard errors across 3 seeds. (\textbf{c}, \textbf{d}) On CIFAR-100, \method\ further improves from combining multiple pre-trained models.}

\end{figure*}

\section{Experiments}
\label{sec:experiments}

We evaluate the proposed method \textbf{Adaptive Feature Transfer (AFT)} across a variety of vision, language, and multi-modal datasets. To probe the effectiveness of the method in the most impactful and practically relevant scenario, we transfer from some of the largest and strongest open-source pre-trained vision and language models such as ViT-G/14 trained with DINOv2~\citep{oquab2023dinov2} and LLaMA-2~\citep{touvron2023llama}. For AFT, we start with a pre-trained version of the downstream architecture and optimize the training loss plus the regularization term in Eq.~\ref{eq:kernel-reg}. We compare AFT against the following methods with comparable computational costs:
\begin{itemize}
    \item \textbf{Standard Transfer Learning (STL)}. STL simply transfers an initialization from the pre-trained model for fine-tuning on the downstream task. This approach prevents the use of any additional pre-trained models that either differ in architecture or size from the downstream model. Therefore we transfer from a pre-trained version of the same downstream architecture with standard fine-tuning. 
    \item \textbf{B-Tuning} \citep{you2022ranking}. In addition to initializing with a pre-trained version of the downstream architecture, B-Tuning uses an approximate posterior predictive distribution of a linear model on top of the features from all other additional pre-trained models as a prior. This method demonstrated state-of-the-art performance when transferring from multiple pre-trained vision models up to ResNet-152~\citep{he2015deep} size. Its effectiveness has yet to be tested for modern massively pre-trained vision foundation models such as Vision Transformers~\citep{ViT}.
    \item \textbf{Knowledge distillation (KD)}. In addition to initializing with a pre-trained version of the downstream architecture, we optimize the feature-based KD objective, which trains the downstream model (student) to fit the pre-trained (teacher) features~\citep{romero2014fitnets}, with the objective given by Eq.~\ref{eq:kd}. We also include two more sophisticated variants of KD, relational knowledge distillation (RKD) \citep{park2019relational}, which aims to capture the relation between the features of different inputs rather than their absolute values, and factor transfer \citep{kim2018paraphrasing}, which trains the student to predict a highly compressed version of the teacher features, where the compression is learned by training an unsupervised autoencoder on the teacher features.
\end{itemize}

All methods start with the same pre-trained initialization of the downstream architecture. \method, B-Tuning, and KD additionally optimize their respective regularization objective weighted by a hyperparameter $\beta > 0,$ which is tuned on the validation set. We will use the term ``pre-trained models" to refer to models whose features $\psi$ are used to define the regularization objectives, rather than being used as the initialization for the downstream model. We include full experiment details, including hyperparameters, in Appendix~\ref{app:train-details}. We report the mean and standard errors computed across 3 runs for each method.

\subsection{Image Classification}
\paragraph{Effective transfer from SOTA vision foundation models.}
We evaluate \method's ability to transfer from state-of-the-art vision foundation models into commonly used downstream architectures, including ViT-S~\citep{ViT}, MLP-Mixer-B~\citep{mlpmixer_b}, and ResNet-50~\citep{he2015deep}. We initialize the downstream models with ImageNet-1K checkpoints for all methods. In Figure~\ref{fig:vision_agg} and ~\ref{fig:vision_all_datasets}, we show performance when transferring from DINOv2 ViT-G/14, the largest model in the DINOv2 family with over a billion parameters, on CIFAR-10~\citep{krizhevsky2009learning}, CIFAR-100~\citep{krizhevsky2009learning}, Oxford Flowers-102~\citep{nilsback2008automated}, Oxford-IIIT Pets~\citep{parkhi2012cats}, \rebut{Describable Textures Dataset (DTD) \citep{dtd} and Food-101 \citep{food-101} datasets.  We find \method\ significantly boosts the performance of all three models, reducing the error by an average of over 15\% relative to STL performance  (Figure \ref{fig:vision_agg}), and outperforms alternatives in most cases.} The main exception is ResNet-50, where KD tends to slightly outperform AFT.

\paragraph{Transfer from multiple pre-trained models} In Figure~\ref{fig:vision_combine_vit} and \ref{fig:vision_combine_mixer}, we show the performance on CIFAR-100 when transferring from various vision foundation models, including BiT ResNet-101x3~\citep{kolesnikov2020big} (denoted BiT), OpenCLIP ViT-G~\citep{cherti2023reproducible, clip} (denoted CLIP) \rebut{and DINOv2 ViT-G/14 \citep{oquab2023dinov2} (denoted DINO)}. \method\ significantly outperforms competing methods. Moreover, \method\ consistently achieves the best performance by transferring from multiple pre-trained models such as DINO + CLIP or BIT + DINO + CLIP. This result shows \method\ can effectively combine complementary features learned by these models due to different inductive biases, pre-training objectives, and pre-training data. 

\begin{figure}[!t]
\centering
    \includegraphics[width=1\linewidth]{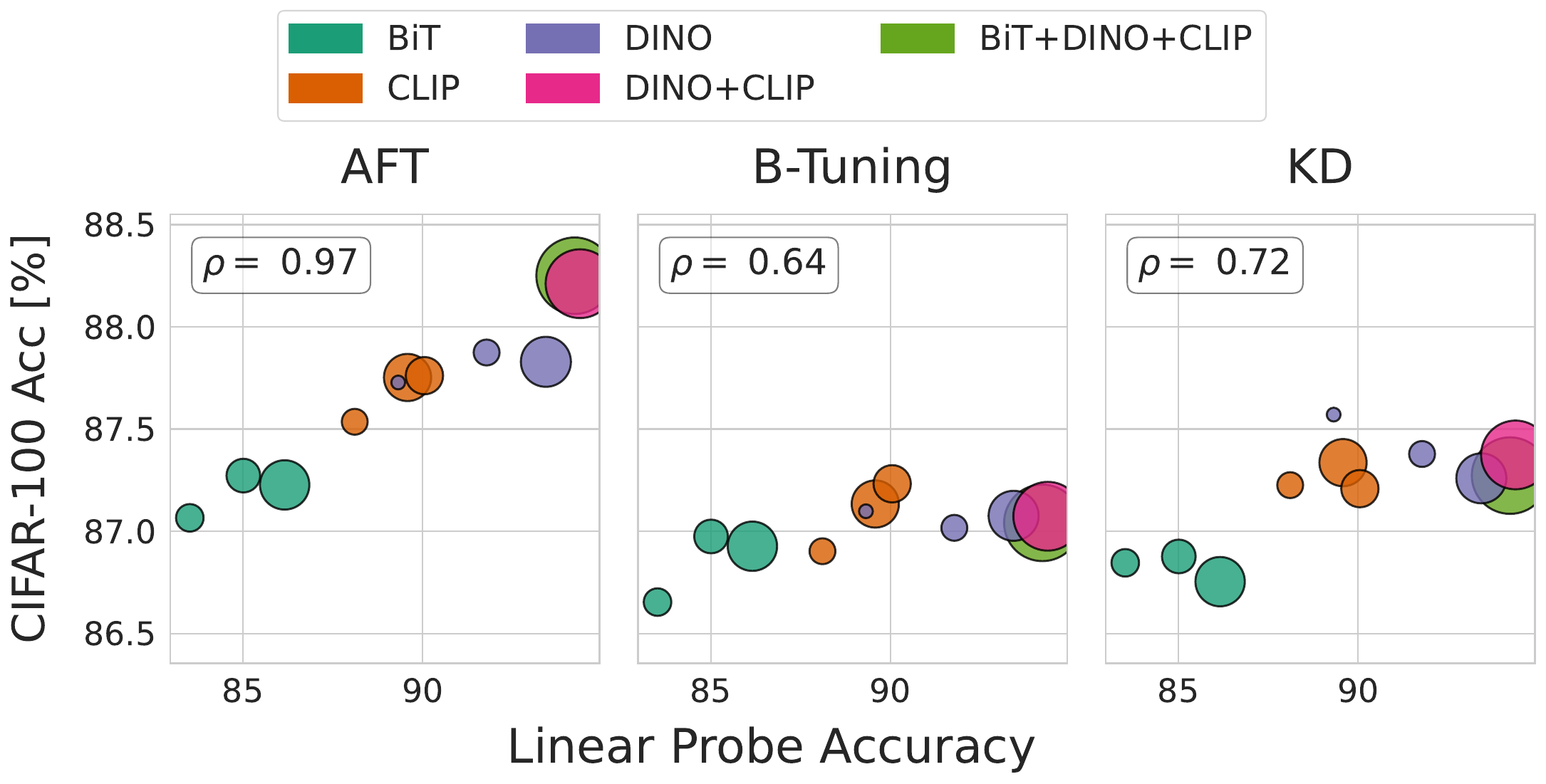}
   \caption{\rebut{\textbf{CIFAR-100 downstream accuracy vs linear probe accuracy of pre-trained features, averaged across 3 downstream models}. \method\ most effectively translates improvements in pre-trained models to improvements in downstream performance. Marker size is proportional to the number of parameters in the pre-trained models, ranging from 87 million to 2.7 billion.
   }}
   \label{fig:vision_corr_aft}
\end{figure}

\paragraph{Performance improves with stronger pre-trained models.}
With an effective method, we wish the downstream performance to consistently improve by transferring from stronger pre-trained models. A method that successfully transfers from large to small models at a particular scale may fail to translate further improvements in pre-trained models to improvements in downstream performance. 

To test the scalability with respect to pre-trained model quality, we compare the downstream performance achieved by each method to the linear probe accuracy of the pre-trained features, i.e., the accuracy achieved by logistic regression on the pre-trained features. We use linear probe accuracy as it measures the amount of useful information we can extract from large pre-trained models on the downstream task without expensive fine-tuning, and is widely used as a metric to estimate the quality of pre-traiend representations as the models are scaled up~\citep{clip, oquab2023dinov2, chen2020generative, ViT}. Figure~\ref{fig:vision_corr_aft} shows \method\ is significantly more effective than alternatives at translating improvements in pre-trained models to improvements in downstream performance, with the highest correlation (0.97) between the downstream accuracy and pre-trained linear probe accuracy. By comparison, \rebut{other methods}' performance saturates early and correlates less well with the linear probe accuracy, showing the unique scalability of \method\ with respect to pre-trained model quality.

\paragraph{Inference time savings.}
\Cref{tab:inference} shows the inference time on CIFAR-100 test set using an NVIDIA A100 GPU for various ViT models. We have shown that AFT effectively transfers from pre-trained models as large as DINOv2 ViT-G/14 to ViT-S/16, which has $50\times$ fewer parameters and $100\times$ faster inference time.

While the linear probe accuracy of a sufficiently large pre-trained model can exceed the accuracy of \method, the linear probe is only efficient to train (via logistic regression) but still expensive to deploy, as it requires inference with the original pre-trained model, and is therefore not a viable alternative to the methods considered here. For example, the linear probe accuracy of OpenCLIP ViT-L/14 roughly matches AFT accuracy when transferred to ViT-S/16 on CIFAR-100 (\Cref{fig:vision_corr_aft}), but OpenCLIP ViT-L/14 is $20\times$ larger than ViT-S/16 and is $4.4\times$ slower to run.

\begin{table}[h!]
\small
\centering
\caption{Inference times on CIFAR-100 test set. Transferring from DINOv2 ViT-G/14 to ViT-S/16 reduces inference times by $100\times$.}
\setlength{\tabcolsep}{8pt}
\renewcommand{\arraystretch}{1.2}
\begin{tabular}{@{}lcc@{}}
\toprule
Model & Params (M) & Inference time (min)\\
\midrule
ViT-S/16 & $22$ & $0.33$\\
OpenCLIP ViT-L/14 & $303$ & $1.45$\\
DINOv2 ViT-G/14 & $1136$ & $34.2$\\
\bottomrule
\end{tabular}
\label{tab:inference}
\end{table}

\begin{figure*}[!t]
\centering
    \includegraphics[width=0.7\linewidth]{figs/legend.pdf}
    \subfloat[Aggregated error]{
    \includegraphics[height=0.2\linewidth]{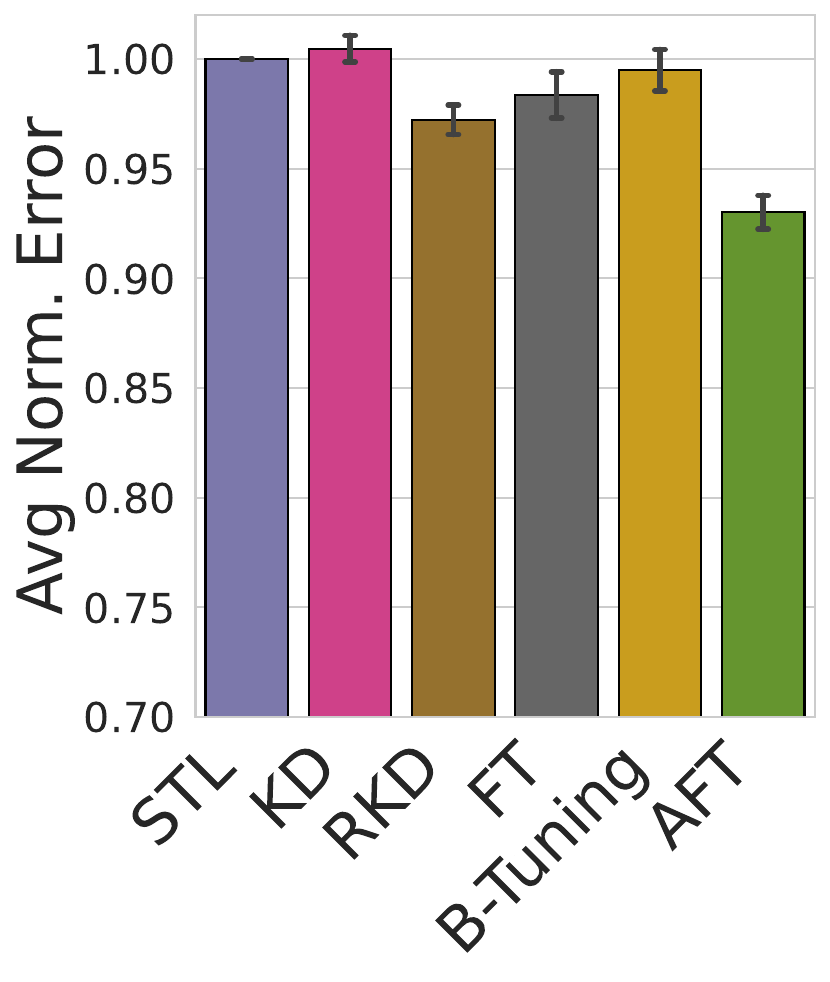}
    \label{fig:nlp_agg}
    }
    \subfloat[Error across models and datasets]{
    \includegraphics[height=0.2\linewidth]{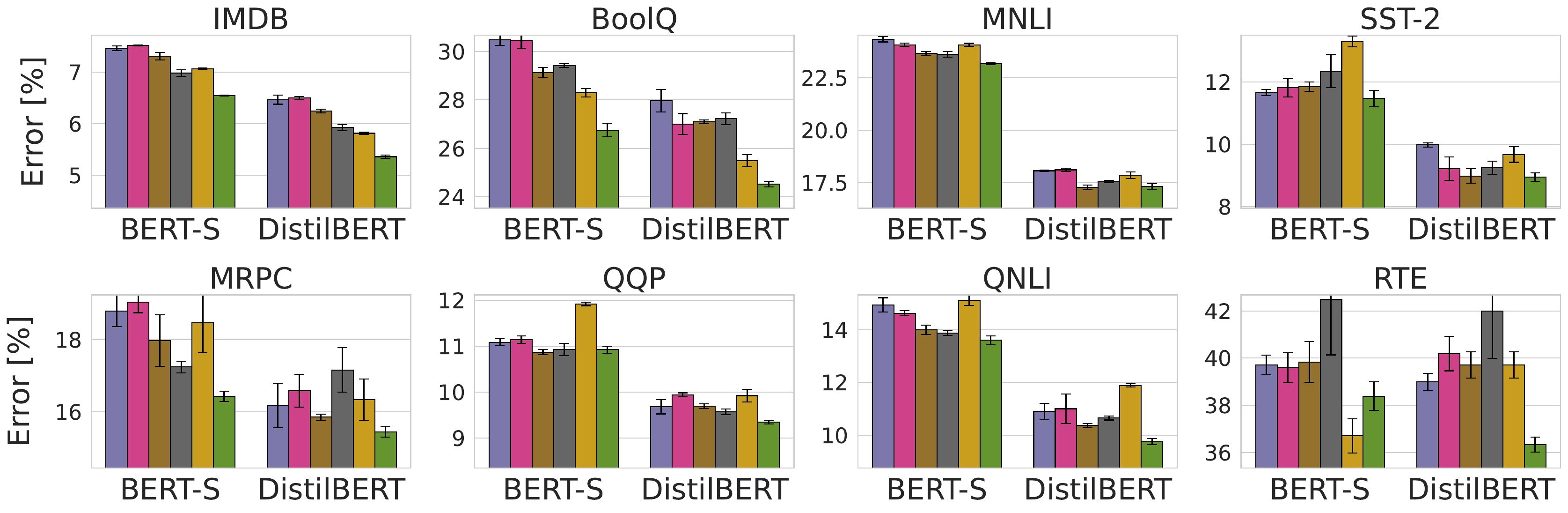}
    \label{fig:nlp_all_datasets}
    }\\
   \caption{\textbf{Evaluation on 8 language dataset using BERT Small and DistillBert as downstream models}. (\textbf{a}) \method\ achieves the lowest normalized error, averaged across 6 datasets, 2 downstream models, and 3 seeds, when transferring from Flan-T5 Large. The error is normalized by the STL error before averaging. The error is normalized by the STL error before averaging. Error bars show standard errors of the aggregated performance. (\textbf{b}) Breakdown of unnormalized error for each downstream model and dataset. Error bars show standard errors across 3 seeds. 
   }

\end{figure*}

\subsection{Natural Language Processing}

\begin{figure}[!t]
\centering
    \includegraphics[width=\linewidth]{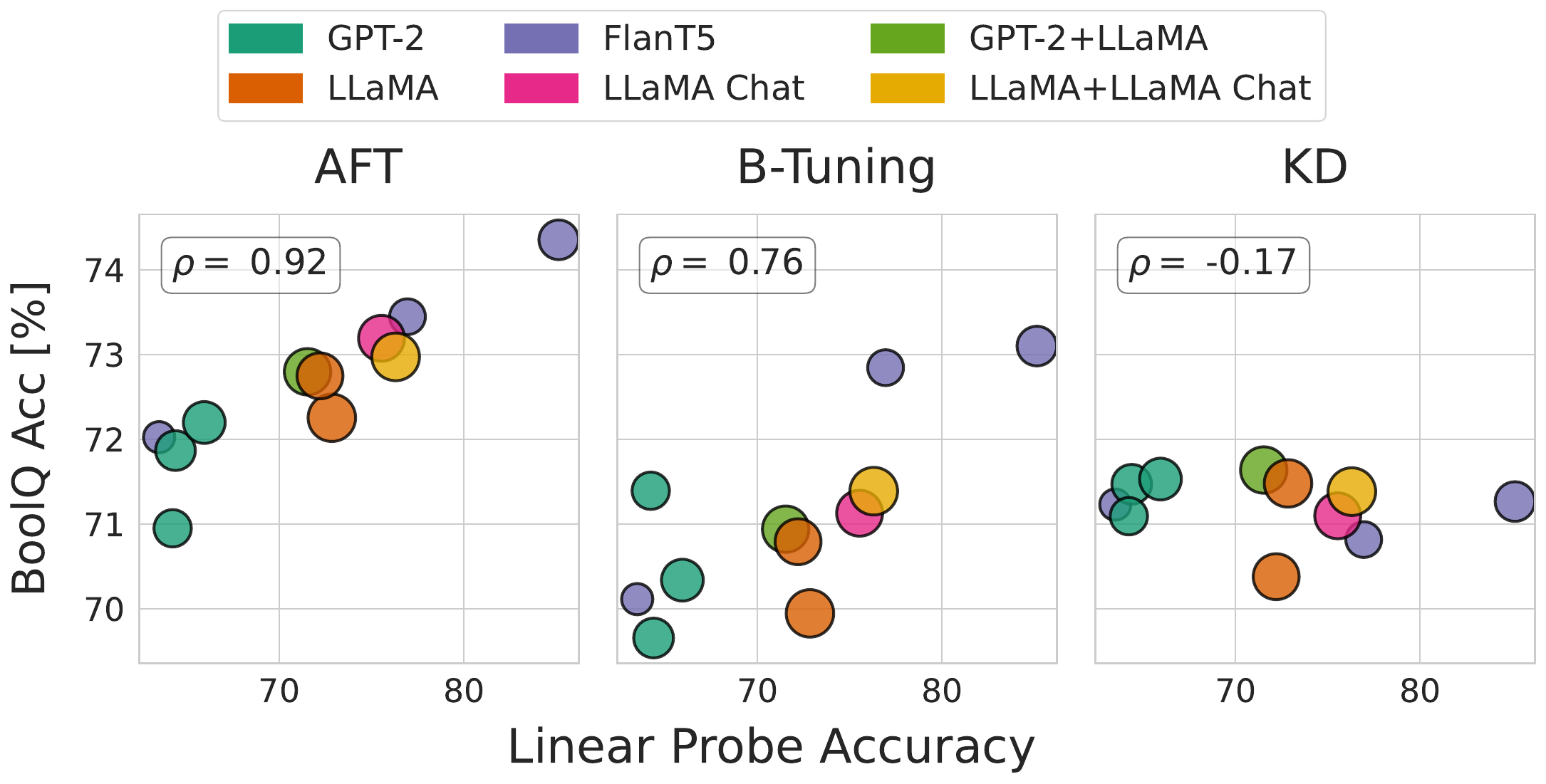}
   \caption{\textbf{BoolQ downstream accuracy v.s. linear probe accuracy of pre-trained features, averaged across two downstream models on BoolQ}.  \method\ most effectively translates improvements in pre-trained models to improvements in downstream performance. Marker size is proportional to the log of the number of parameters in the pre-trained models, ranging from 61 million to 14 billion.
   }
   \label{fig:nlp_corr_aft}
\end{figure}

We explore transferring from larger open-source large language models, such as GPT-2~\citep{gpt_2}, Flan-T5~\citep{flan_t5}, and LLaMA 2~\citep{touvron2023llama}, into much smaller language models, namely BERT Small~\citep{Bert} and DistillBERT~\citep{distilbert}. We follow common practices for extracting input-level features: using the embedding of the [CLS] token for BERT models and the decoder's embedding of the last token for GPT-2, Flan-T5, and LLaMA. In \Cref{app:langauge}, we provide details on input formatting and discuss memorization concerns.

\rebut{We evaluate the performance of AFT and competing methods at transferring from Flan-T5 Large to BERT Small and DistillBERT on 8 datasets: Large Movie Review (IMDB)\citep{maas-EtAl:2011:ACL-HLT2011}, BoolQ \citep{wang2019superglue}, MNLI \citep{MNLI}, SST-2 \citep{sst-2}, MRPC \citep{MRPC}, QQP \citep{glue}, QNLI \citep{QNLI}, and RTE \cite{glue}. In \Cref{fig:nlp_agg,fig:nlp_all_datasets}, we show that AFT significantly outperforms the competing methods.} As in the vision datasets, \method\ most effectively translates improvements in pre-trained models to improvements in downstream performance. In \Cref{fig:nlp_corr_aft}, we observe that using \method\ with instruction-tuned pre-trained language models like Flan-T5 and LLaMA Chat leads to the best post-transfer performance, aligning with their superior zero-shot question answering capabilities \citep{flan_t5}.

In \Cref{fig:nlp_corr_aft}, unlike in vision datasets, we find that combining multiple pre-trained models often does not improve their linear probe accuracy or the accuracy achieved by \method, suggesting little complementary information is learned between these pre-trained language models. This may be due to the high similarity in pre-training datasets, objectives, and architectures among these transformer-based generative models, which are predominantly trained with next or masked token prediction on similar distributions of internet text.

\subsection{Multi-modality} 
\method's ability to efficiently transfer from multiple models makes it well-suited for multi-modal applications. In these settings, modality-specific sub-components, such as image and text encoders in CLIP~\citep{clip}, can benefit from transferring complementary features learned by pre-trained models in each modality. We demonstrate this on SNLI-VE \citep{xie2019visual, xie2018visual}, a visual entailment dataset where the goal is to determine if a text corresponds to an image. Using ResNet-50 CLIP as the downstream model, we construct a classifier $f_\theta(x_I, x_T) = W \phi(x_I, x_T)$ with features $\phi(x_I, x_T)$ given by the tensor product $\phi_I(x_I) \otimes \phi_T(x_T)$, representing pairwise interactions between image and text features. Table~\ref{tab:snli} shows that \method\ improves CLIP's performance by simultaneously transferring from a ViT-L/14 trained with DINOv2 and LLaMA 13B.

\begin{table}[h!]
\small
\centering
\caption{\method\ improves CLIP's accuracy on SNLI-VE by transferring from DINOv2 and LLaMA 13B.}
\begin{tabular}{llll}
\toprule
 Method & STL & \makecell{KD} & \makecell{\method}\\
\midrule
SNLI-VE Acc. & $73.69_{\pm0.28}$ & $74.05_{\pm0.05}$ & $\mathbf{74.39_{\pm0.18}}$\\
\bottomrule
\end{tabular}
\label{tab:snli}
\end{table}

\begin{figure*}[!t]
\centering
    \subfloat[AFT upweights informative features]{
    \includegraphics[height=0.2\linewidth]{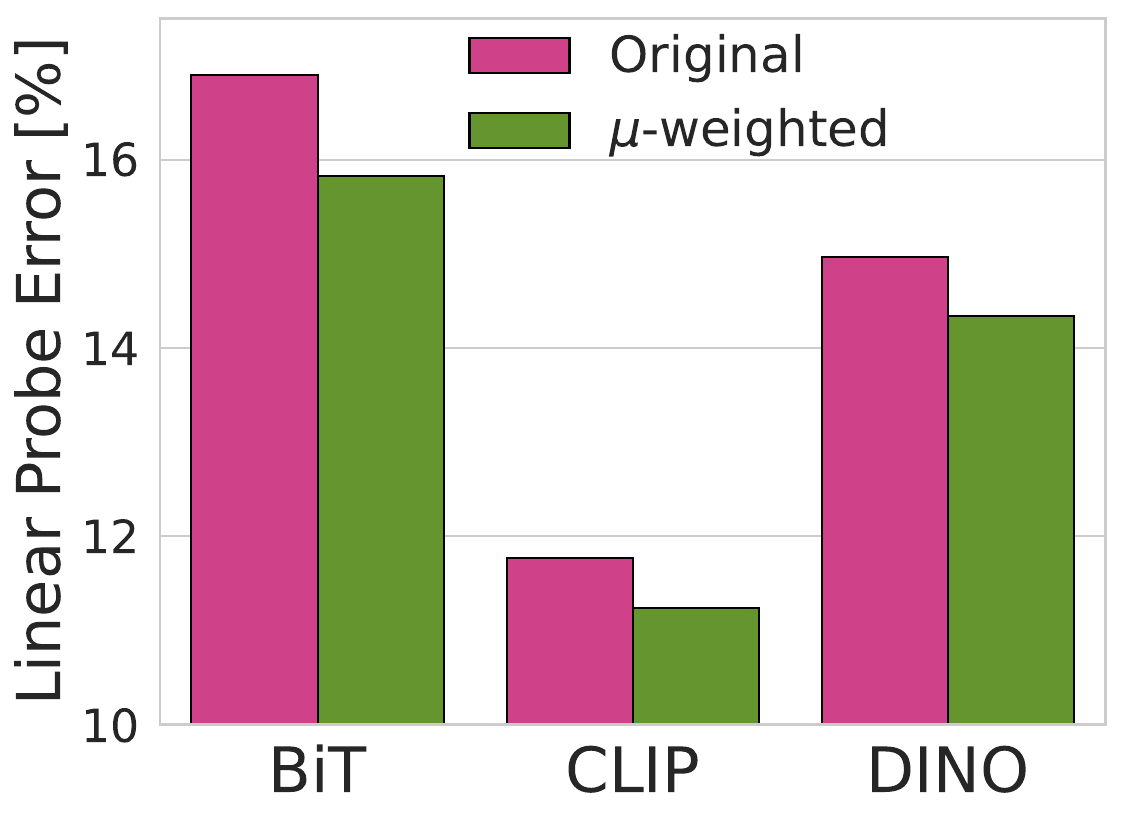}
    \label{fig:weighted_acc}
    }
    \subfloat[Error v.s. $d_\mathrm{noise}$]{
    \includegraphics[height=0.2\linewidth]{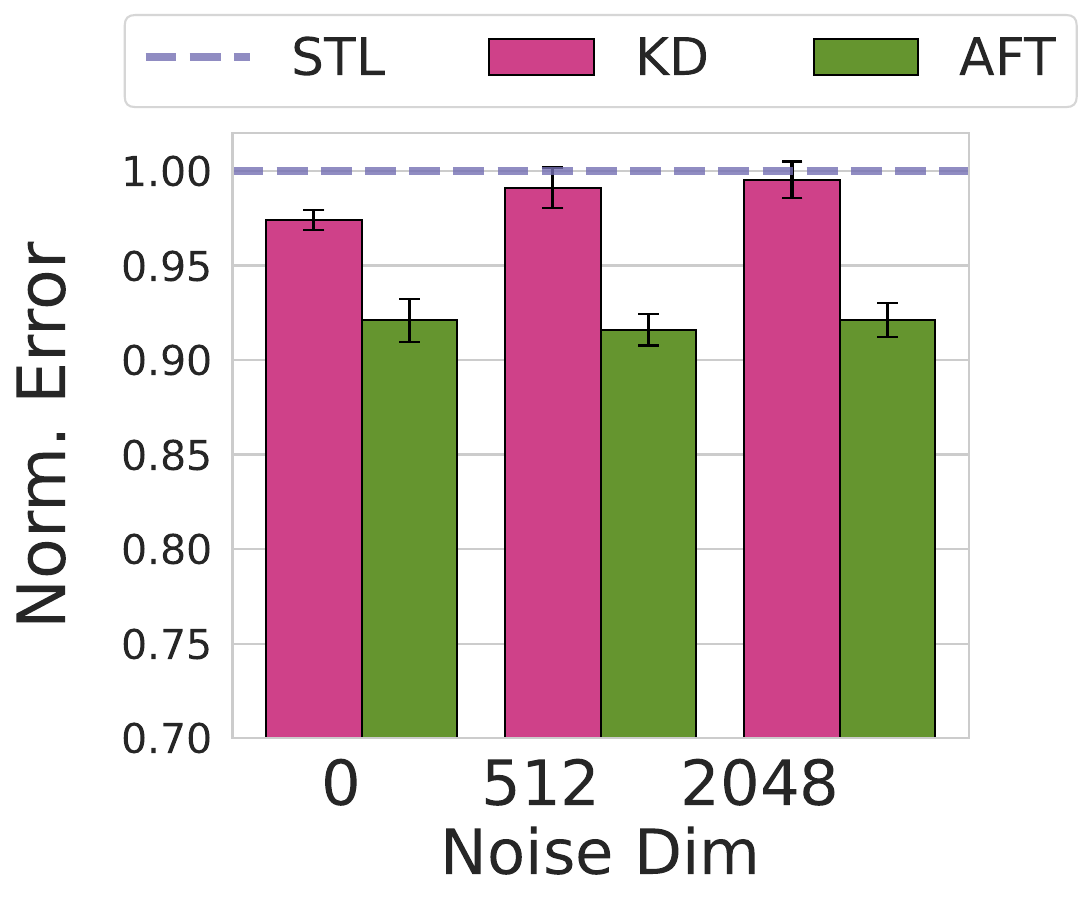}
    \label{fig:noise_acc}
    }
    \subfloat[Distribution of $\mu_i$]{
    \includegraphics[height=0.2\linewidth]{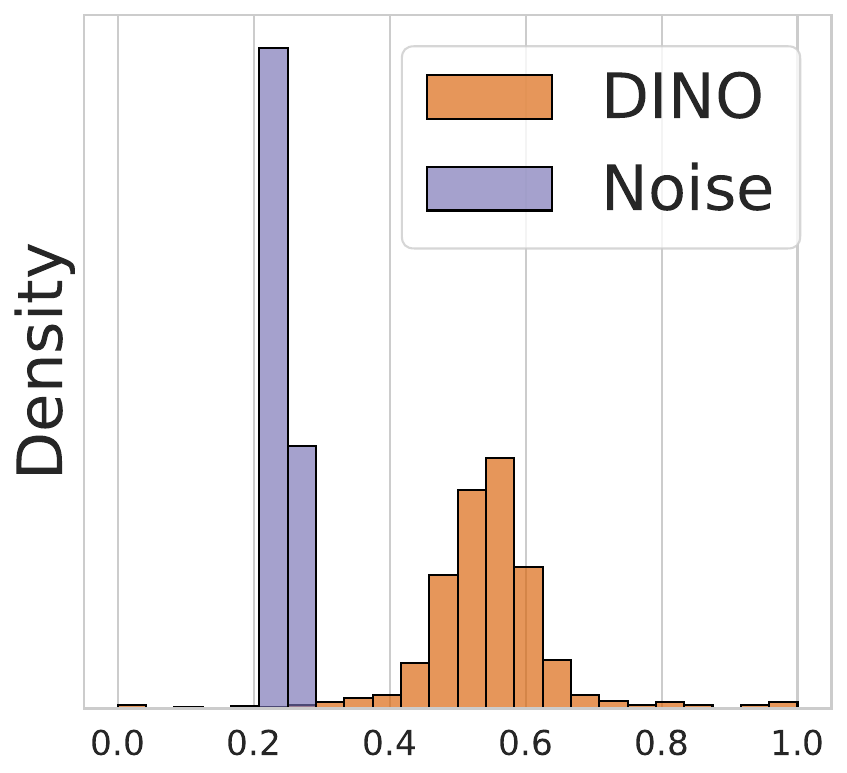}
    \label{fig:noise_rho}
    }
    \caption{\textbf{Analysis of \method's properties on CIFAR-100}. (\textbf{a}) Linear probe error is improved when applying the learned AFT weights $\mu$ to the pre-trained features, indicating that AFT effectively upweights informative features for the downstream task. (\textbf{b}) \method's performance remains stable as an increasing number of noise features ($d_\mathrm{noise}$) are appended to the pre-trained features, demonstrating its robustness to uninformative features. (\textbf{c}) The learned $\mu_i$ values effectively separate noise features from useful features, with noise features assigned much smaller weights.}
   \label{fig:analysis}
\end{figure*}

\section{Analyzing Why AFT works}
Having demonstrated AFT as a highly effective method, we now perform experiments to verify our understanding of why AFT works and reveal which design decisions are important.

\subsection{\method\ upweights features that generalize better}
If the learned weights $\mu$ in AFT indeed upweight the more informative features, then we expect a linear probe trained on the weighted features $\mu \psi$ should outperform one trained on the original features $\psi.$
In \Cref{fig:weighted_acc}, we show the linear probe error on CIFAR-100 with the original pre-trained features $\psi$ from BiT 50x3, OpenCLIP ViT-G, or DINOv2 ViT-G, and on the weighted features $\mu \psi$, where the weights $\mu$ are learned by AFT when transferring to ViT-S. We find weighing the pre-trained features by the AFT weights improves the linear probe performance for all pre-trained models, showing that AFT indeed identifies and upweights pre-trained features that leads to better generalization on the downstream task.

\subsection{\method\ is robust to uninformative features}
\label{sec:viz_rho}
As the adaptive nature of \method\ enables it to automatically downweight irrelevant features without any intervention, we expect it to perform well even when a large number of pre-trained features are completely uninformative of the downstream task. To test this hypothesis, we transfer from DINOv2 ViT-G/14 and a random noise model whose features are drawn from $\N(0, I_{d_\mathrm{noise}}),$ where $d_\mathrm{noise} \in \{0, 512, 2048\}$ is its feature dimension, into ViT-S/16 on CIFAR-100.

Results in Figure~\ref{fig:noise_acc} clearly illustrate the limitations of compression-based objectives like KD, whose performance quickly degrades to near STL level as we introduce the noise features, since the downstream model is trained to learn many useless features. By constrast, \method\ performance is nearly unaffected by the presence of noise features. In Figure~\ref{fig:noise_rho}, we show this robustness because the learned weights $\mu_i$ in \method\ are much smaller for the noise features.

\subsection{Ablation experiments}
\label{sec:ablation}
We investigate the impact of key design choices in \method\ on its performance on CIFAR-100 and BoolQ. We compare \method\ with \rebut{four other} variants where we a) do not use the kernel formulation and instead use the $\ell_2$ objective in Eq.~\ref{eq:l2-reg} (No kernel), b) disable the ability to learn $\mu$ and fix it to be the identity (Identity $\mu$), \rebut{c) Use a dense rather than diagonal $\mu$ (Dense $\mu$), d) replace the linear kernel $k(x,x') = \phi(x)^\top \phi(x')$ with radial basis function (RBF) kernel $k(x,x') = \exp(-\norm{\phi(x) - \phi(x')}^2)$ (RBF), and e) use bi-level optimization over $\theta$ and $\mu$ by performing 5 inner updates for $\mu$ per update of $\theta$ (Bi-level).} 

\rebut{We find using the kernel formulation and learning the feature weights $\mu$ are essential to \method's performance, while the use of alternative kernels such as the RBF kernel and bi-level optimization does not impact the performance in any significant way. Learning a dense rather than diagonal $\mu$ slightly hurts performance.}

\section{Discussion}

Transfer learning --- pre-training then fine-tuning --- is becoming the mainstream paradigm for deploying deep learning models. However, the default approach to transfer learning remains surprisingly naive, transferring limited and generic information: simply use the pre-trained weights as an initialization for the downstream loss optimization. There is therefore a great need to develop transfer learning procedures more tailored to the task at hand. 

Through AFT, we have shown that a simple, general, and computationally efficient approach exists for transferring knowledge from large models to small models. An important takeaway from AFT is that aligning what is transferred to the small downstream model with the specific downstream task is crucial for effective transfer learning, showing this large-to-small transfer fundamentally differs from just model compression. As future works uncover even more effective methods for large-to-small transfer, our fundamental understanding of transfer learning will further advance.

\method\ offers a trade-off between reducing the cost of transfer learning and the potential performance improvements. \method\ is inherently limited by the reduced representational capacity of small downstream models. This limitation can be mitigated by selecting more expressive downstream models, albeit at the cost of diminished savings in training and inference. Furthermore, the current formulation of \method\ prioritizes simplicity, generality, and computational efficiency by restricting the transfer to only the last layer features. Expanding and optimizing the set of features transferred via \method\ is an exciting direction for future work that may significantly further enhance performance.

\section*{Acknowledgements} We thank Micah Goldblum, Nate Gruver, and Daohan Lu for helpful discussions. This work is supported by NSF CAREER IIS-2145492,
NSF CDS\&E-MSS 2134216, NSF HDR-2118310, BigHat Biosciences, Capital One, and an Amazon Research Award.

\section*{Impact Statement}
The goal of this work is to advance the field of Machine Learning. There are many potential societal consequences of our work, none which we feel must be specifically highlighted here.

\bibliography{refs}
\bibliographystyle{iclr2024_conference}

\clearpage

\appendix
\onecolumn
\section{Experiment details} \label{app:train-details}
We tune the hyperparameter $\beta$ for \method, KD, and B-Tuning in all experiments by holding out 10\% of the original training set and selecting the $\beta$ value that yields the highest accuracy on this holdout set. Once the optimal $\beta$ is determined, we train the models on the entire training set using this value. Our implementations of relational knowledge distillation (RKD) and B-Tuning are based on their original implementations, available at \url{https://github.com/lenscloth/RKD} and \url{https://github.com/thuml/LogME}, respectively. Following \citet{park2019relational}, we weigh the angle loss and the distance loss in RKD at a 2:1 ratio. For Factor Transfer, we replace the original CNN-based paraphraser and translator networks with MLPs, as we work with the last layer features, which lack spatial dimensions, instead of the intermediate CNN feature maps used in the original paper \citep{kim2018paraphrasing}.

\subsection{Vision experiments}
We use the timm~\citep{rw2019timm} implementation for all vision models, their pre-trained checkpoints, and data preprocessing pipelines. We do not use data augmentation in any experiment. 

We use the Adam optimizer in all experiments and train for 5000 steps (rounded up to whole epochs) with a batch size of 128 and a cosine lr decay schedule. We use a base learning rate of $1e-4$ for ViT-S/16 and MLP Mixer-B, and $1e-3$ for ResNet-50. We tune $\beta \in \{3, 10, 30\}$ for \method, $\beta \in \{0.1, 1, 10, 100\}$ for KD, RKD, FT, \rebut{and $\beta \in \{1, 1e2, 1e3, 1e4\}$ for B-Tuning}. We use the Adam optimizer and a learning rate of $1e-2$ for updating the vector $s$ parameterizing the diagonal elements of $\mu.$

\subsection{Language experiments} \label{app:langauge}
We use the Hugging Face implementation of all the language models. We use the Adam optimizer in all experiments and train for 5000 steps (rounded up to whole epochs) with a batch size of 64 and a cosine lr decay schedule. We use a base learning rate of $2e-5$ for both BERT Small and DistilBERT. We tune $\beta \in \{1, 3, 10\}$ for \method, $\beta \in \{0.01, 0.1, 1, 10\}$ for KD, RKD, FT,  \rebut{and $\beta \in \{1, 1e2, 1e3, 1e4\}$ for B-Tuning}. We use the Adam optimizer and a learning rate of $1e-2$ for updating the vector $s$ parameterizing the diagonal elements of $\mu.$

We format each example as follows before feeding it into the language model:
\begin{itemize}
\item IMDB \citep{maas-EtAl:2011:ACL-HLT2011}: ⟨review⟩ Overall, the sentiment of my review is
\item BoolQ \citep{wang2019superglue}: Question: ⟨question⟩\textbackslash n Reference: ⟨passage⟩\textbackslash n Answer:
\item MNLI \citep{MNLI}: Premise: ⟨premise⟩\textbackslash n Hypothesis: ⟨hypothesis⟩\textbackslash n Does the premise entail the hypothesis? Answer:
\item SST-2 \citep{sst-2}: Review: "⟨sentence⟩"\textbackslash n Sentiment:
\item MRPC \citep{MRPC}: Sentence 1: ⟨sentence1⟩\textbackslash n Sentence 2: ⟨sentence2⟩\textbackslash n Is Sentence 1 equivalent to Sentence 2? Answer:
\item QQP \citep{glue}: Question 1: ⟨question1⟩\textbackslash n Question 2: ⟨question2⟩\textbackslash n Are Question 1 and Question 2 equivalent? Answer:
\item QNLI \citep{QNLI}: Question: ⟨question⟩\textbackslash n Sentence: ⟨sentence⟩\textbackslash n Does the sentence answer the question? Answer:
\item RTE \citep{glue}: Sentence 1: ⟨sentence1⟩\textbackslash n Sentence 2: ⟨sentence2⟩\textbackslash n Does Sentence 1 entail Sentence 2? Answer:
\end{itemize}

\paragraph{On memorization concerns.}
Language models are pre-trained on internet-scale data, making it difficult to rule out the possibility that the benchmarks we evaluated on are not in their training set. However, this concern is irrelevant for us as our experiments aim only to compare each method's effectiveness in transferring knowledge from the pre-trained models rather than establishing some absolute level of downstream performance on these benchmarks.

\subsection{SNLI-VE experiments}
We use the official OpenAI implementation of CLIP ResNet-50 \citep{clip}. We use the Adam optimizer in all experiments and train for 1 epoch with a batch size of 64. We use a base learning rate of $1e-5$ for CLIP ResNet-50. We tune $\beta \in \{1, 3, 10\}$ for \method, and $\beta \in \{0.01, 0.1, 1\}$ for KD. We use the Adam optimizer and a learning rate of $1e-2$ for updating the vector $s$ parameterizing the diagonal elements of $\mu.$

\section{\rebut{Extended results}}

\begin{table}[h!]
\centering
\caption{Unnormalized results for transfer to ViT-S/16 in Figure~\ref{fig:vision_combine_vit}.}
\label{tab:results_vits}
\begin{tabular}{lccccc}
\toprule
Method & BiT & CLIP & DINO & DINO+CLIP & BiT+DINO+CLIP \\
\midrule
KD & $87.79_{\pm0.07}$ & $88.06_{\pm0.06}$ & $88.17_{\pm0.06}$ & $87.96_{\pm0.21}$ & $88.13_{\pm0.01}$ \\
B-Tuning & $88.01_{\pm0.05}$ & $88.57_{\pm0.06}$ & $88.54_{\pm0.11}$ & $88.66_{\pm0.13}$ & $88.67_{\pm0.04}$ \\
AFT & $88.25_{\pm0.09}$ & $88.56_{\pm0.06}$ & $88.88_{\pm0.06}$ & $89.23_{\pm0.10}$ & $89.14_{\pm0.00}$ \\
\bottomrule
\end{tabular}
\end{table}

\begin{table}[h!]
\centering
\caption{Unnormalized results for transfer to MLP-Mixer in Figure~\ref{fig:vision_combine_mixer}.}
\label{tab:results}
\begin{tabular}{lccccc}
\toprule
Method & BiT & CLIP & DINO & DINO+CLIP & BiT+DINO+CLIP \\
\midrule
KD & $86.21_{\pm0.05}$ & $86.63_{\pm0.13}$ & $86.42_{\pm0.11}$ & $86.55_{\pm0.27}$ & $86.40_{\pm0.06}$ \\
B-Tuning & $87.34_{\pm0.06}$ & $87.42_{\pm0.10}$ & $87.20_{\pm0.16}$ & $87.43_{\pm0.02}$ & $87.27_{\pm0.04}$ \\
AFT & $87.40_{\pm0.03}$ & $87.92_{\pm0.02}$ & $87.76_{\pm0.11}$ & $88.23_{\pm0.07}$ & $88.42_{\pm0.02}$ \\
\bottomrule
\end{tabular}
\end{table}

\begin{table}[h!]
\centering
\caption{Unnormalized results for transfer to ResNet-50.}
\label{tab:results_rn50}
\begin{tabular}{lccccc}
\toprule
Method & BiT & CLIP & DINO & DINO+CLIP & BiT+DINO+CLIP \\
\midrule
KD & $86.64_{\pm0.15}$ & $87.32_{\pm0.16}$ & $87.18_{\pm0.10}$ & $87.62_{\pm0.07}$ & $87.29_{\pm0.14}$ \\
B-Tuning & $85.57_{\pm0.10}$ & $85.42_{\pm0.04}$ & $85.49_{\pm\text{NaN}}$ & $85.06_{\pm0.05}$ & $85.19_{\pm0.11}$ \\
AFT & $86.17_{\pm0.05}$ & $86.78_{\pm0.07}$ & $86.91_{\pm0.09}$ & $87.18_{\pm0.04}$ & $87.08_{\pm0.10}$ \\
\bottomrule
\end{tabular}
\end{table}

\end{document}